\documentclass[sigconf,authorversion,nonacm]{acmart} 

\usepackage{utils}

\usepackage{cleveref}

\crefname{algocf}{alg.}{algs.}
\Crefname{algocf}{Algorithm}{Algorithms}

\usepackage{special/PSalgorithm}

\usepackage{special/notationTable}

\usepackage{special/resultTableStyle}

\usepackage{mathtools}

\usepackage{graphicx}
\usepackage{subcaption}

\begin{document}

%%
%% The "title" command has an optional parameter,
%% allowing the author to define a "short title" to be used in page headers.
\title{Mining Potentially Explanatory Patterns via Partial Solutions}
\subtitle{A proposed methodology and experimental results}

%%OLD TITLE
%Towards Explaining Genetic Algorithms through Partial Solutions

%%ALTERNATIVE TITLE
%Mining Potentially Explanatory Patterns via Partial Solutions

%%
%% The "author" command and its associated commands are used to define
%% the authors and their affiliations.
%% Of note is the shared affiliation of the first two authors, and the
%% "authornote" and "authornotemark" commands
%% used to denote shared contribution to the research.
\author{GianCarlo A.P.I. Catalano}
\email{g.a.catalano@stir.ac.uk}

\author{Alexander E.I. Brownlee}
\email{alexander.brownlee@stir.ac.uk}

\author{David Cairns}
\email{dec@cs.stir.ac.uk}
\affiliation{
  \institution{University of Stirling}
  \department{Computing Science and Mathematics}
  \city{Stirling}
  \country{UK}
}

\author{John McCall}
\email{j.mccall@rgu.ac.uk}
\affiliation{
  \institution{Robert Gordon University}
  \department{National Subsea Centre}
  \city{Aberdeen}
  \country{UK}
}

\author{Russell Ainslie}
\email{russell.ainslie@bt.com}
\affiliation{
  \institution{BT Technology}
  \department{Applied Research}
  \city{Ipswich}
  \country{UK}
}

%% Since it's unclear what to call the basis 
\newcommand{\PSbasis}{catalog\ }

%%
%% By default, the full list of authors will be used in the page
%% headers. Often, this list is too long, and will overlap
%% other information printed in the page headers. This command allows
%% the author to define a more concise list
%% of authors' names for this purpose.
\renewcommand{\shortauthors}{Catalano et al.}

%%
%% The abstract is a short summary of the work to be presented in the
%% article.
\begin{abstract}
   Genetic Algorithms have established their capability for solving many complex optimization problems. Even as good solutions are produced, the user's understanding of a problem is not necessarily improved, which can lead to a lack of confidence in the results. To mitigate this issue, explainability aims to give insight to the user by presenting them with the knowledge obtained by the algorithm.

 In this paper we introduce Partial Solutions in order to improve the explainability of solutions to combinatorial optimization problems. Partial Solutions represent beneficial traits found by analyzing a population, and are presented to the user for explainability, but also provide an explicit model from which new solutions can be generated.
 We present an algorithm that assembles a collection of Partial Solutions chosen to strike a balance between high fitness, simplicity and atomicity.
Experiments with standard benchmarks show that the proposed algorithm is able to find Partial Solutions which improve explainability at reasonable computational cost without affecting search performance.
\end{abstract}


%% To save space
% \ccsdesc[500]{Computing methodologies~Search methodologies}

%%
%% Keywords. The author(s) should pick words that accurately describe
%% the work being presented. Separate the keywords with commas.

%%
%% This command processes the author and affiliation and title
%% information and builds the first part of the formatted document.
\maketitle

\section{Introduction}

Genetic Algorithms (GA) are able to solve many kinds of optimization problems by producing high-quality results, which promotes their adoption even in critical applications \cite{XAISurvey, AIExceedHuman}. A problem presented by GAs, but also by Machine Learning, is that they act as black boxes which humans struggle to trust, despite the apparent quality of the outputs \cite{XAIConcepts}.  This mistrust is not unjustified: a GA's results can be sub-optimal for many reasons, especially due to stochastic behavior, incorrect fitness function implementations, unintended biases and design issues \cite{XAISurvey}.

This is not a small problem, since user trust becomes increasingly important as GAs are introduced into high-stake applications \cite{XAISurvey}. 
The solution to this issue is \textbf{explainability}, where the user is presented with the knowledge synthesized by the algorithm to solve the task. This can be used to debug the implementation, provide documentation for auditing purposes, manually adjusting and improve understanding of computer-generated solutions \cite{XAISurvey}.

\subsection{Optimization problems}
The present paper will concentrate on combinatorial optimization problems, which  can be particularly complex to solve despite their relatively simple construction.
Many have been shown to be computationally hard, such as Graph Coloring \cite{GuideToNP}, Knapsack \cite{KnapsackProblems} and resource allocation problems \cite{CombinatorialOptimization}.
As such, these often have to be approached using metaheuristics such as GAs, but their internal mechanisms are often difficult for humans to follow, suggesting the need for techniques similar to XAI for Machine Learning \cite{XAISurvey}.

There are many open questions which apply to optimization problems, that are also particularly suited for explainability:
\begin{itemize}
    \item \textbf{Why is a solution good, or better than another?}
    \item \textbf{What characterizes good solutions?}
    \item \textbf{How can the global optima be detected}?
\end{itemize}

The approach we propose offers one way to address these questions. We introduce the concept of a \textit{Partial Solution} (PS), an explicit decomposition of a positive trait identified in high-fitness solutions. A collection of PSs is assembled and acts as a model that is then sampled to generate new solutions. In doing so, the algorithm is inherently explainable, offering the ability for the end user to identify key components of the recommended solutions.

These Partial Solutions offer a novel way of providing both global and local explanations, but at the same time improve the search efficiency when solving the problem. 
We test whether it is possible to consistently find these Partial Solutions for some benchmark functions which have known underlying structures, in \Cref{subsubsec:T1} and \Cref{subsubsec:T2}. Afterwards, the ability to solve problems efficiently is compared against a traditional GA and UMDA in \Cref{subsubsec:T3}.

\section{Related work}
We now summarize relevant research in the EC and optimization literature related to explainability. %, and give an overview of related research topics.

\subsection{Models within GAs and EDAs}
Approaches which use a model to represent the problem, including surrogate fitness functions such as neural networks and decision trees, or the probabilistic models of Estimation of Distribution Algorithms (EDA), have been a popular target for developing explainability functionality. This is because these models are designed to represent abstract information in order to accurately generalize the solution space. This knowledge can also be presented to the user for explainability purposes \cite{IntersectionEvoCom}, drawing on the techniques developed for explainability in Machine Learning.

This paper makes use of related concepts from EDAs, as they present a fertile avenue for explainability applications that can be adapted to metaheuristics.
Although traditional GAs lack explicit models, such structures can be added. For example, \citet{MarkovNetoworkValueAdded} approximates the fitness function using a Markov Network, which can be then presented to the user, whereas \citet{InterpretableFeatureSelectionManifacturing} use a marginal product model for linkage learning, and \citet{ConqueringHierarchicalDifficulty} use a hierarchical linkage model.
Some methods can be argued to be using the population itself as a model, such as \cite{NDSXAITrajectory, PCATrajectoryMining}, where the GA is analyzed in terms of the \textbf{trajectory} within the search space that the population takes during the evolution process. Similar to how an EDA  generates its model from a population, the present work will analyze a population and produce a set of PSs, which then act as a model from which new solutions are generated. On the other hand, since this ``model'' is only built once and is not iterative, our approach is not a fully fledged EDA.
\Cref{sec:FormalDefinition} will give a formal definition of the proposed concepts, and \Cref{sec:Novelty} will review the novelty of this approach in more detail.

\subsection{Backbones and building blocks}
As discussed in \cite{IntersectionEvoCom}, \textbf{backbones} are related to explainability because they give insight into patterns found in solutions. Backbones mostly relate to SAT problems, and consist of sub-configurations of values which are shared among the global optima.
Mathematically, backbones and PSs share some similarities because they both represent commonalities within sets of solutions, but PSs are made with explainability in mind, and thus they are generated using different quality metrics and used very differently.

Both concepts are reminiscent of \textbf{schemata}, introduced by Holland in some early theoretical work to formalize the convergence properties of  GAs \cite{HollandAdaptation, OverviewSchemaTheory}, but these theories were later shown to not match observed results \cite{BuildingBlockFallacy}.
It should be noted that these results are limited to Holland’s conjectures, and do not relate to the explicit construction of Partial Solutions or their use in explainability, which will be the central topics of this paper.

%Partial Solutions can in principle also be extended to higher order relationships between variables than schemata.

\section{Partial Solutions}
\subsection{Informal Introduction}
\label{subsec:InformationIntroduction}
To understand the behavior of the fitness function of a combinatorial optimization problem, one can look at patterns associated with higher fitness solutions. 
These may be higher-order interactions between groups of variables, e.g. `all variables in the group must be the same value', but in the present paper the focus is on patterns where certain variables are fixed and others can vary (indicated using \PS{*}).
For example \PS{0**1} applies to \PS{0001}, \PS{0011}, \texttt{0101} and \PS{0111}.

\begin{figure}[ht]
  \centering
  \includegraphics[width=\linewidth]{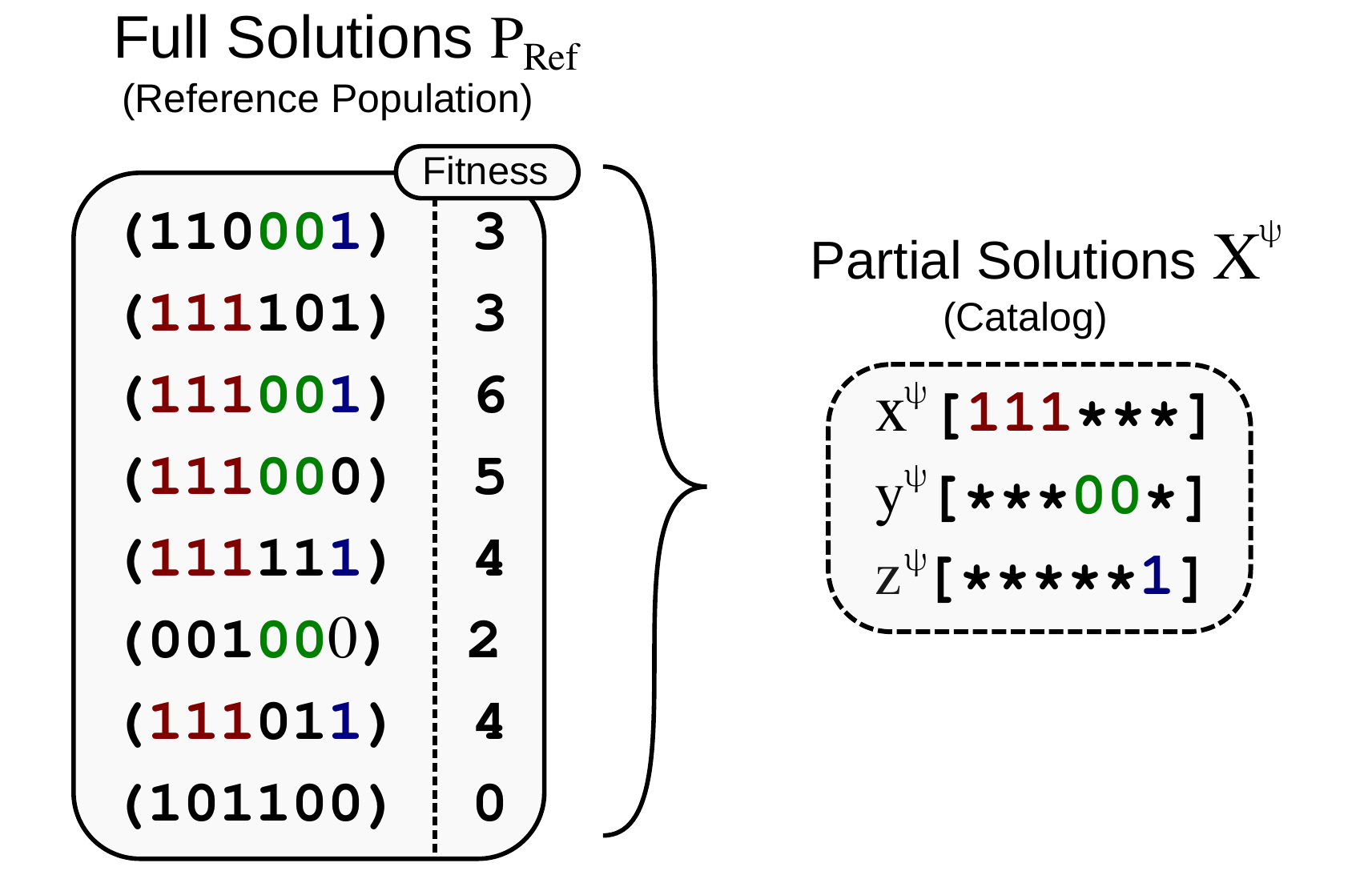}
  \caption{The positive traits in a collection of full solutions $\text{P}_\text{Ref}$\ can be described by a collection of Partial Solutions $X^\psi$}
\label{fig:PSCompression} 
  \Description{The items in $X^\psi$ represent beneficial traits that were observed in the population $X$. While $X_3$ is the global optimum, the rest of the population contains small imperfections which allow the algorithm to determine what makes $X_3$ good.}
\end{figure}

A collection of such patterns, found by analyzing a population, provides a succinct and expressive representation of what makes a solution ``good'' (as seen in \Cref{fig:PSCompression}).

In a real problem, PSs can offer useful insight into what variables are linked, and what values they should take.
In an employee-allocation problem, for instance, these sub-configurations might indicate that it is beneficial to assign certain groups of employees to specific tasks.
A solution such as \PS{1111111} might have its high fitness explained by the presence of \PS{111***} and \PS{*****1}, although \PS{111001} is even better because of the presence of \PS{***00*}.

Additionally, PSs might show the presence of positive traits that cannot coexist (\textit{e.g.}, \PS{**00} and \PS{*11*}), which can explain the presence of many near-global optima. The ability to find such disagreeing PSs can help in explaining situations where generally positive traits are not necessarily found in the global optima.

%% INSUFFICIENTSPACE
% \begin{equation*}   %% I'm not allowed to mess with \vspace!!!!!!
%    \PS{***111111**}
% \end{equation*}
% \begin{equation*}
%     \PS{*000*******}
% \end{equation*}
% \begin{equation*}
%     \PS{*******0000}
% \end{equation*}
% \begin{equation*}
%     \PS{111********}
% \end{equation*}

Partial solutions are an explicit decomposition of the problem, which has two benefits:
\begin{itemize}
    \item Partial Solutions are \textbf{interpretable}:
    \begin{itemize}
        \item \textbf{Global explanations}: describing the fitness landscape by finding simple and atomic Partial Solutions which are associated with high-fitness.
        \item \textbf{Local explanations}: describing a solution by pointing out the Partial Solutions it contains.
    \end{itemize}
    \item Partial Solutions \textbf{assist the search process}, because they act as a model that describes high-fitness regions of the solution space (similar to Probabilistic Graphical Models) from which solutions may be constructed (see \Cref{sec:GeneratingFullSolutions}).
\end{itemize}

\subsection{Formal definition}
\label{sec:FormalDefinition}

\begin{table}[ht]
  \caption{Notation}
  \label{tab:notation}
  \begin{tabular}{cl}
    \toprule
    Symbol &Meaning\\
    \midrule
    $F$ & Fitness function to be maximized\\
    $\Fpsi$ & Fitness function for partial solutions\\
    $n$ & The number of parameters in the solutions\\
    $a, b, \dots,x$ & Variables used to denote a \emph{full} solution \\
    $x_1, x_2, \dots,x_n$ & The parameters of $x$, a full solution\\
    $a^\psi, b^\psi, \dots,\xpsi$ & Variable used to denote a \emph{Partial} Solution \\
    $\xpsi_1, \xpsi_2, \dots,\xpsi_n$ & The parameters of $\xpsi$, a \textit{Partial Solution} \\
    $\ReferencePopulation$ & A collection of evaluated full solutions\\

  \bottomrule
\end{tabular}
\end{table}

\Cref{tab:notation} summarizes our notation.
A Partial Solution $\xpsi$ can be described as ``\textit{a sub-configuration of parameter values which is associated with high-fitness solutions within a reference population}'', where that reference population will be denoted as  $\ReferencePopulation$.

In the same way that a full solution is a tuple of values $x = (x_1, x_2,...,x_2)$, a Partial Solution is $\xpsi = (\xpsi_1, \xpsi_2, ... \xpsi_n)$ where each $\xpsi_i$ can take any value that $x_i$ can, as well as ``\PS{*}'', the ``any'' value.
A solution $x$ is said to ``contain'' the Partial Solution $\xpsi$ when they agree on the fixed values found in $\xpsi$.
\begin{equation}
\text{contains}(x, \xpsi) \Leftrightarrow (\xpsi_i = *) \vee (\xpsi_i = x_i)\ \forall\ i \in [1, n]
\end{equation}

Conversely, it is possible to find all the solutions in $\ReferencePopulation$ which contain the Partial Solution $\xpsi$:
\begin{equation}
    \text{observations}_{\ReferencePopulation}(\xpsi) = \{x \in \ReferencePopulation\ |\ \text{contains}(x, \xpsi) \}
    \label{eq:Observations}
\end{equation}
For an optimization problem, within all of the possible PSs there is a small subset which is preferable for explainability, the \textbf{\PSbasis}. These are the PSs which are optimal with respect to three metrics described in the next section: \textit{simplicity}, \textit{mean fitness} and \textit{atomicity}.

\subsection{Partial Solution metrics}
It is preferable for a PS to have many of its parameters take the \PS{*} value, since fewer fixed parameters improve interpretability. We define this metric to be:
\begin{equation}
\text{simplicity}(\xpsi) = |\{ i\ \in\ [1, n] |\ \xpsi_i = *\}|
\end{equation}

The simplest PS possible has no fixed variables, and consists entirely out of \PS{*}'s.
Other than being simple, a PS also needs to be associated with high-fitness solutions, which can be quantified by finding the mean fitness of its observations (cf. \Cref{eq:Observations}). 
\begin{equation}
    \text{meanFitness}(\xpsi) = \text{average}(F(\text{observations}_{\ReferencePopulation}(\xpsi))\
\end{equation}

In order to have PSs assist with explainability, the previous two metrics are not sufficient. Assume $\xpsi$ and $\ypsi$ are optimal in different metrics, and let $\zpsi$ be the union of their fixed values.
$\zpsi$ is likely to also be high-scoring, since it will have similar observations to its parents, and its \texttt{simplicity} might not be significantly worse than $\xpsi$ and $\ypsi$, even though it is redundant compared to them.

This effect was described in Holland’s Building Block Hypothesis as beneficial to the GA process, but for our purposes this is actually undesirable, since it gets in the way of interpretability due to all the redundancy present.
To solve this problem, it will be necessary to define a third metric, which will encapsulate the idea that a PS should be “irreducible”: \textbf{atomicity}.

% \subsection{Atomicity}
For a Partial Solution $\xpsi $, its fixed-value elements $ \xpsi_{j} $ each contribute in some way to its average fitness. We define a normalized fitness function, remapped in the range [0, 1]:

\begin{align}
    \FNorm(x) = \frac{\text{F}(x) - m}{S}  \\\nonumber
    (\text{where}\ m = \min_{x\ \in\ \ReferencePopulation} F(x), S = &\sum_{x\ \in\ \ReferencePopulation} F(x) - m) 
\end{align}
    
In order to measure the contribution of the variable $\xpsi_k$ within $\xpsi$, define the following modifications of $\xpsi$:
\begin{itemize}
    \item \textbf{isolate}: where only $\xpsi_k$ is fixed
    \item \textbf{exclude}: where $\xpsi_k$ is \PS{*}
\end{itemize}
And we define their ``benefit'' to be the sum of the normalized fitnesses of their observations.

\begin{align}
    &\text{isolate}(\xpsi, k) = \ypsi\ s.t.\ \ypsi_i = \left\{\begin{matrix}
\xpsi_i & \text{if}\ i = k\\ 
* & \text{otherwise} 
\end{matrix}\right. \\
    &\text{exclude}(\xpsi, k) = \zpsi\ s.t.\ \zpsi_i = \left\{\begin{matrix}
\xpsi_i & \text{if}\ i \neq k\\ 
* & \text{otherwise} 
\end{matrix}\right. \\
& \text{benefit}(a^\psi) = \sum_{x\ \in\ \text{observations}_{\ReferencePopulation}(a^\psi)} F_{\text{Norm}(x)} 
\end{align}

Finally, we define the contribution of the variable $\xpsi_k$ as:
\begin{align}
    \text{contribution}(\xpsi, k) & = p_{AB}\ \cdot\ \log\left ( \frac{p_{AB}}{p_{A*}\ \cdot\ p_{*B}} \right ) \\\nonumber
    \text{where}\ p_{AB} & = \text{benefit}(\xpsi), \\\nonumber
    p_{A*} &= \text{benefit}(\text{isolate}(\xpsi, k)), \\\nonumber 
     p_{*B} &= \text{benefit}(\text{exclude}(\xpsi, k)) 
\end{align}

This is derived from mutual information, as used in DSMGA-II \cite{DSMGAII}.
Continuing the example from \Cref{subsec:InformationIntroduction}, the Partial Solution \PS{111**1} will have its contributions calculated at $k \in \{1, 2, 3, 6\}$, and since independent groups are present the contribution scores will be low. For \PS{111***}, the contributions are calculated at $k \in \{1, 2, 3\}$, which will all be greater than those in the previous example.

In the same way that a chain is as strong as its weakest link, the atomicity of $\xpsi$ is defined as the \textit{minimum} of its contributions, which ensures that all of the fixed parameters in the PS depend on each other and are essential.
\begin{equation}
    \text{atomicity}(\xpsi) = 
\text{min}\left\{\left. \text{contribution}(\xpsi, k))\ \right|\substack{\ k\ \in\ [1,\ n] \\ \ \xpsi_k \neq\ \texttt{*}}\ \right\}
\end{equation}

Each of these metrics is necessary in order to characterize the PSs which are best suited for explainability:
\begin{itemize}
    \item \textbf{Simplicity} prioritizes more understandable PSs
    \item \textbf{Mean-fitness} promotes beneficial PSs
    \item \textbf{Atomicity} prevents redundant PSs
\end{itemize}

\subsection{Novelty}
\label{sec:Novelty}
Having defined Partial Solutions, we can now review related existing concepts to demonstrate where our novelty lies.

\textbf{Schemata} and \textbf{backbones} are similarly structured to Partial Solutions, but they are used very differently. \textbf{Schemata} are abstractions that were used for theoretical work on GAs, and are \textit{implicitly sampled}, i.e. not meant to be explicitly instantiated. In contrast, our work does \textbf{instantiate} them, similar to other studies which use them to improve search efficiency \cite{NaturalApproachSchemaProcessing, DiscoveringDeepBuildingBlocks}, whereas in this work they are designed mainly for explainability.

\textbf{Backbones} relate to SAT problems where multiple satisfactory solutions share a common sub-configuration. For most problems there is only one backbone, and it is preferred for it to have many fixed variables in order to shrink the search space \cite{BackbonesInOptimization}, whereas in this work the aims are reversed to improve interpretability.

Since in this work the PS \PSbasis is treated as a model, EDAs are also an adjacent topic. In this sense, there are some related works:
\textbf{Linkage learning} \cite{OptimalDependencyTrees, DSMGAII} and \textbf{Family-of-Subsets} \cite{GOMEA} aim to find which variables are interacting in the fitness function, and are then assembled into structures such as sets or cliques \cite{SurveyOptimizationBuilding}.
These structures only determine which variables are interacting, but their optimal value assignments are not specified, whereas PSs contain both. 
In other words, PSs are instances of the cliques / linked sets found by the aforementioned approaches.

\section{Overall system}
\label{sec:OverallSystem}
The system that we propose consists of:
\begin{itemize}
    \item \textbf{PS Miner}: obtains the PS \PSbasis from a reference population, described in \Cref{subsec:Algorithm} and \Cref{algo:PSMiner}
    \item \textbf{Pick \& Merge}: an algorithm which forms  solutions by combining elements from the \PSbasis, described in \Cref{sec:GeneratingFullSolutions} and \Cref{algo:PandM}.
\end{itemize}

The following (one pass) sequence of operations combines them:
\begin{enumerate}
    \item Generate a reference population $\ReferencePopulation$, and evaluate it
    \item Apply the PS Miner on $\ReferencePopulation$ to generate the PS \PSbasis
    \item Apply \texttt{pick\_and\_merge} on the PS \PSbasis to obtain full solutions
\end{enumerate}

\begin{figure}[!ht]
  \centering
  \includegraphics[width=\linewidth]{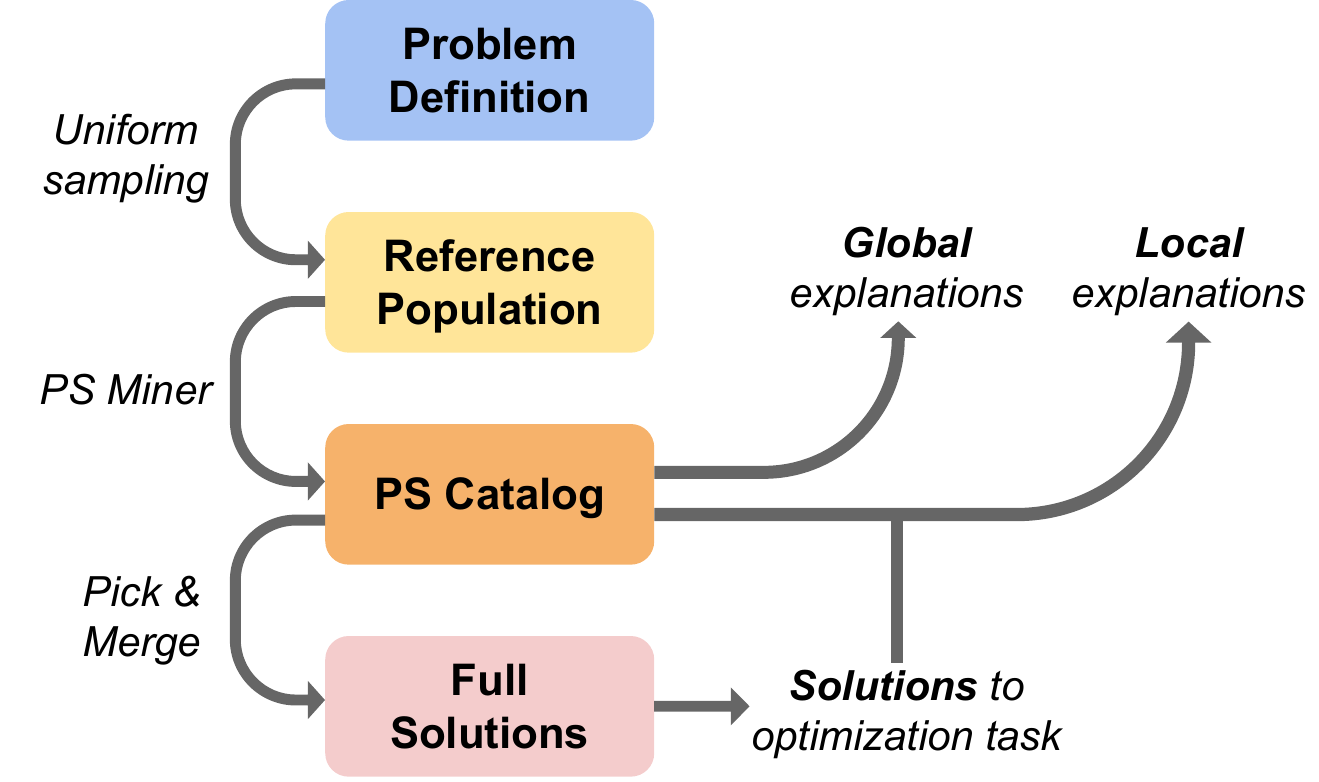}
  \caption{The proposed system, which uses the PS \PSbasis for both explainability and to generate full solutions.}
  \label{fig:OurSystem}
   \Description{By using Partial Solutions as an intermediate, the system is able to generate full solutions and explainability, specifically local explanations (when comparing against full solutions) and global explanations.}
\end{figure}

The process of producing solutions offers the explainability benefits of PSs and maintains the ability to solve the original optimization problem (see \Cref{fig:OurSystem}), all of which will be tested in \Cref{sec:Experiments}.

\subsection{Finding Partial Solution Catalogs}
\label{subsec:Algorithm}
The algorithm to find the PS \PSbasis will be implemented using the metrics in \Cref{sec:FormalDefinition}, but the design of the search method involves some non-trivial design choices, which we will now describe.

\subsubsection{Reference Population}
$\ReferencePopulation$ is used to evaluate the \texttt{meanFitness} and \texttt{atomicity} of PSs, and remains fixed throughout the PS Mining process. It's important for $\ReferencePopulation$ to cover enough of the search space, and in the present work this was done using random uniform sampling. \Cref{subsubsec:T2} investigates the necessary size of $\ReferencePopulation$ and whether it might be beneficial to evolve it. 

\begin{table}[hb]
\centering
\caption{PS local search variants}
\label{tab:LocalSearchVariants}
\scriptsize
\begin{tabular}{lrrr}\toprule
\textbf{Variant} &\thead{\textbf{\texttt{get\_local}}} &\thead{\textbf{\texttt{get\_init}}} \\\midrule
\textbf{Simplification-only} &\makecell[c]{Simplifications of $\xpsi$} & \makecell{returns the top individuals in \\ $\ReferencePopulation$ as  Partial Solutions}  \\\midrule
\textbf{Specialization-only} &\makecell[c]{Specializations of $\xpsi$ }& \makecell{returns the universal PS: **...**} \\\midrule
\textbf{Full-local} & \makecell[c]{$\text{simplifications} $\\ $\cup$ \\ $\text{specializations}$} &The union of the above two methods \\
\bottomrule
\end{tabular}
\end{table}
\subsubsection{Multiple objectives and $\Fpsi$}
To simplify the system,\textbf{ the three objectives are aggregated} into a single objective $\Fpsi$. This is obtained by remapping each metric into the range $[0, 1]$ and taking the mean, as seen in \Cref{algo:PSMiner}.

Note that an evaluation of a full solution using $F$ is a different operation from evaluating a Partial Solution with $\Fpsi$, and the latter uses pre-calculated fitness values from $\ReferencePopulation$ for efficiency.

\subsubsection{Exclusion archive}
A problem to be tackled when searching for the \PSbasis is that the PS search space is large and contains many local optima clusters, which hinder the search process. Their presence is due to the fact that local optima for $F$ also act as fully-set PS local optima for \texttt{meanFitness}, and their neighborhoods contain many similar individuals but with `\PS{*}' values.
In order to assist the algorithm, an \textbf{exclusion archive} based method is proposed, where the items selected in every iteration are put into an archive and are not allowed to reappear in the population again. In another sense, it marks which regions have been visited and prevents the population from revisiting them.
The archive stores all of the best solutions seen throughout the iterations of the algorithm, and upon termination these are evaluated and the top scoring PSs are returned. This approach is implemented as shown in \Cref{algo:PSMiner}, and it was tested as part of \Cref{subsubsec:T1}.

\newenvironment{grayBox}[1]{
    \begin{tcolorbox}[colback=gray!10, colframe=black, left=2pt, right=2pt, top=2pt, bottom=2pt]
            #1
}{
    \setstretch{1.0} % resets the interline spacing
    \end{tcolorbox}
}

\begin{algorithm}
    \caption{Archive-based PS Miner}
    \label{algo:PSMiner}
    \DontPrintSemicolon

    \SetKwFunction{FRemap}{\texttt{remap}}
    \SetKwFunction{FFpsi}{$F^\psi$}
    \SetKwFunction{FTop}{\texttt{top}}
    \SetKwFunction{FSelect}{\texttt{select}}
    \SetKwFunction{FGetInit}{\texttt{get\_init}}
    \SetKwFunction{FGetLocal}{\texttt{get\_local}}
    \newcommand{\vpopsize}{\texttt{pop\_size}}
    \newcommand{\vqtyret}{\texttt{qty\_ret}}
    
    \SetKwFunction{FMinePS}{\texttt{mine\_ps}}

    \SetKwProg{Fn}{Def}{:}{}

    \Fn{\FRemap{values}}{
        \KwRet $\frac{values-\min(values)}{\max(values)-\min(values)}$
    }

    \Fn{\FFpsi{$\Xpsi$}}{
        $M \leftarrow \FRemap (\texttt{meanFitness}(\Xpsi))$ \;
        $S \leftarrow \FRemap (\texttt{simplicity}(\Xpsi))$ \;
        $A \leftarrow \FRemap (\texttt{atomicity}(\Xpsi))$ \;
        \KwRet $\{\texttt{avg}(m, s, a)\ \texttt{for}\ m, s, a \in \texttt{zip}(M, S, A)\}$
    }

    \Fn{\FTop{$\Xpsi$, \texttt{quantity}}}{
        $\text{sorted} \leftarrow \text{sort}(\Xpsi, \text{key} = \FFpsi)$ \;
        \KwRet \texttt{sorted[:}quantity\texttt{]}
    }

    \begin{grayBox}
        \FGetInit and \FGetLocal are discussed in \Cref{subsec:Algorithm}
        
    \end{grayBox}

  \Fn{\FMinePS{\FGetInit, \FGetLocal, \vpopsize, \vqtyret}}{
    \newcommand{\vPSpop}{\Xpsi}
    \newcommand{\vPSarchive}{\text{archive}}
    \newcommand{\vPSselected}{\text{selected}}
    $\vPSpop \leftarrow \FGetInit{}$ \;
    $\vPSarchive \leftarrow \{\} $ \;

    \While{\texttt{termination\_criteria\_not\_met}}{
        $\vPSselected \leftarrow \texttt{TournamentSelection}(\vPSpop)$ \;
        
        $\text{localities} \leftarrow \bigcup_{\xpsi \in\ \text{seleted}}\text{get\_local}(\xpsi)$ \;

        $\vPSarchive \leftarrow \vPSarchive \cup \text{selected}$ \;
        
        $\vPSpop \leftarrow (\vPSpop \cup \text{localities}) \smallsetminus \vPSarchive$ \;

        $\vPSpop \leftarrow \FTop(\vPSpop, \vpopsize)$ \; 
    }

    \KwRet $\FTop(\psArchive, \vqtyret)$ \;
  }
\end{algorithm}
\begin{algorithm}
    \caption{Pick-and-Merge algorithm}
    \label{algo:PandM}
    \DontPrintSemicolon

    \SetKwFunction{FWeightedRandomChoice}{\texttt{weighted\_random\_choice}}
    \SetKwFunction{FMergeFrom}{\texttt{merge\_from}}
    \SetKwFunction{FFillGaps}{\texttt{fill\_gaps}}
    \SetKwFunction{FGenerate}{\texttt{generate\_via\_pick\_and\_merge}}
    \newcommand{\vlimit}{\texttt{limit}}
    \newcommand{\vavailable}{\texttt{available}}
    \newcommand{\vadded}{\texttt{added}}
    
    \SetKwProg{Fn}{Def}{:}{}

    \Fn{\FWeightedRandomChoice{$\Xpsi$}}{
        \tcc{samples a random element of $\Xpsi$ using $F^\psi$ as weights}
    }

    \Fn{\FMergeFrom{$\Xpsi$, $\vlimit$}}{
        $\vavailable \leftarrow \texttt{copy}(\Xpsi)$ \;
        $\xpsi \leftarrow \PS{**..**}$ \;
        $\vadded \leftarrow 0$ \;
        \While{$\vavailable \neq \varnothing$ \& $\vadded < \vlimit$}{
            $\ypsi \leftarrow \FWeightedRandomChoice{\vavailable}$ \;
            $\vavailable \leftarrow \vavailable \smallsetminus \{\ypsi\}$ \;
            \If{$\texttt{mergeable}(\xpsi,\ypsi)$} 
            {
            $\xpsi \leftarrow \texttt{merge}(\xpsi,\ypsi)$ \;
            $\vadded \leftarrow \vadded + 1$ \;
            }
        }
        \KwRet $\xpsi$ \;
    }

    \Fn{\FFillGaps{$\xpsi$}}{
        \For{$\xpsi_i \in \xpsi$}{
            \If{$\xpsi = *$}{
                $\xpsi_i \leftarrow \texttt{random}.\texttt{randrange}(\texttt{cardinalities}[i])$ \;
            }
        }
        \KwRet $\xpsi$
    }

    \Fn{\FGenerate{$\Xpsi$, \texttt{merge\_limit}}}{
        \KwRet \FFillGaps{\FMergeFrom{$\Xpsi$, \texttt{merge\_limit}} } \;
    }
\end{algorithm}

%% INSUFFICIENTSPACE
%\begin{figure}[ht]
%  \label{fig:FitnessLanscape}
%  \centering
%  \includegraphics[width=\linewidth] {images/fitness_landscape.pdf}
%  \caption{A PS fitness landscape for a problem with two global optima, which correspond to clusters of local optima in the PS landscape.}
%   \Description{In a problem with two global optima, \PS{11111} and \PS{00000}, the Partial Solutions will be clustered around their corresponding Partial Solutions. In this example, each of the clusters has many local optima, and only the basis Partial Solutions are the true global optima.}
% \end{figure}

\subsubsection{Local search}
Knowing that the PS search space might contain many clustered optima, it is beneficial to include a form of local search.
Given a PS $\xpsi$, its local neighborhood consists of 
\begin{itemize}
    \item \textbf{Simplifications}: one of the fixed variables is replaced by \PS{*}
    \item \textbf{Specializations}: one of the \PS{*} is replaced by a fixed value
\end{itemize}

Local search may use either or both of these, which determine the implementation of \texttt{get\_local} and \texttt{get\_init} in \Cref{algo:PSMiner}. The three possible implementations will be tested as part of \Cref{subsubsec:T1}.

\subsection{Pick \& Merge algorithm}
\label{sec:GeneratingFullSolutions}
The PS \PSbasis obtained via \Cref{algo:PSMiner} can be used as a model to construct high-quality full solutions.
\Cref{algo:PandM} simply  picks items from the PS \PSbasis and merges them when possible. At the end, any remaining unfixed parameters are filled with random values, so that a full solution can be returned.
Let $\Xpsi$ be the \PSbasis found for a problem, and define the following operations, which are used in \Cref{algo:PandM}:

\begin{align}
    \text{mergeable}(\xpsi, \ypsi) & = \neg \exists\ i\ s.t.\ (\xpsi_i \neq * \neq \ypsi_i \wedge \xpsi \neq \ypsi) \\
    \text{merge}(\xpsi, \ypsi) & = \zpsi\ s.t.\ \zpsi_i = \left\{\begin{matrix}
\xpsi_i & \text{if}\ \xpsi_i \neq *\\ 
\ypsi_i & \text{if}\ \ypsi_i \neq *\\ 
* & \text{otherwise}
\end{matrix}\right. \\
\text{to\_full}(\xpsi) &= x\ s.t.\ x_i = \xpsi_i\ \forall\ i \in [1, n] \\\nonumber
 & (\text {defined only when}\ \neg \exists\ \xpsi_i =\ *)
\end{align}

The parameter \texttt{merge\_limit} used in \pickAndMerge\ determines how many PSs should be contained in a generated solution, which for this paper was set to $\lceil \sqrt{n} \rceil$.
Note how this algorithm does not use the original fitness function, as it is assumed that $\Xpsi$ is representative of beneficial traits.

\section{Experiments}
\label{sec:Experiments}

This section will present three research questions, \textbf{RQ1}, \textbf{RQ2} and \textbf{RQ3}, answered by the testing rounds \textbf{T1}, \textbf{T2} and \textbf{T3} respectively.

\begin{itemize}
    \item \textbf{RQ1}: Which algorithm parameters are best suited for finding the PS \PSbasis?
    \item \textbf{RQ2}: How is the ability to extract Partial Solutions from a reference population affected by its size and  the amount of generations it has been evolved for?
    \item \textbf{RQ3}: Can Partial Solutions be used to obtain high-fitness full solutions? Our goal is increasing explainability, but the algorithm must still be able to solve the problem.
\end{itemize}

T1 and T2 will check which algorithms can produce the PS \PSbasis effectively, which we assume will help explainability. For testing, we'll use benchmark problems with known catalogs, defined in \Cref{subsec:BenchmarkProblems}.
T3 will test the system's ability to solve the original optimization problem by checking how often it is able to find the global optima.
The algorithms and the parameters being tested will be contained in \Cref{subsec:Setup}, along with the description of the testing rounds. Finally, the results are presented and discussed in  \Cref{sec:Results}.

\subsection{Benchmark Problems}
\label{subsec:BenchmarkProblems}
Here we draw upon benchmarks previously used by the EDA community, as they have clearly defined structures that map well to graphical models, and present well-defined ``\textbf{target}'' PSs which we aim for our algorithms to find.
The benchmark problems will need to be challenging enough to benefit from explainability, and hence ``simpler'' ones such as OneMax and BinVal will not be considered. 
Instead, problems which have dependencies between variables will be chosen, and to prevent any biases due to proximity in linked variables, the input variables will have their positions shuffled.

\begin{figure}[ht]
  \centering
  \includegraphics[width=\linewidth]{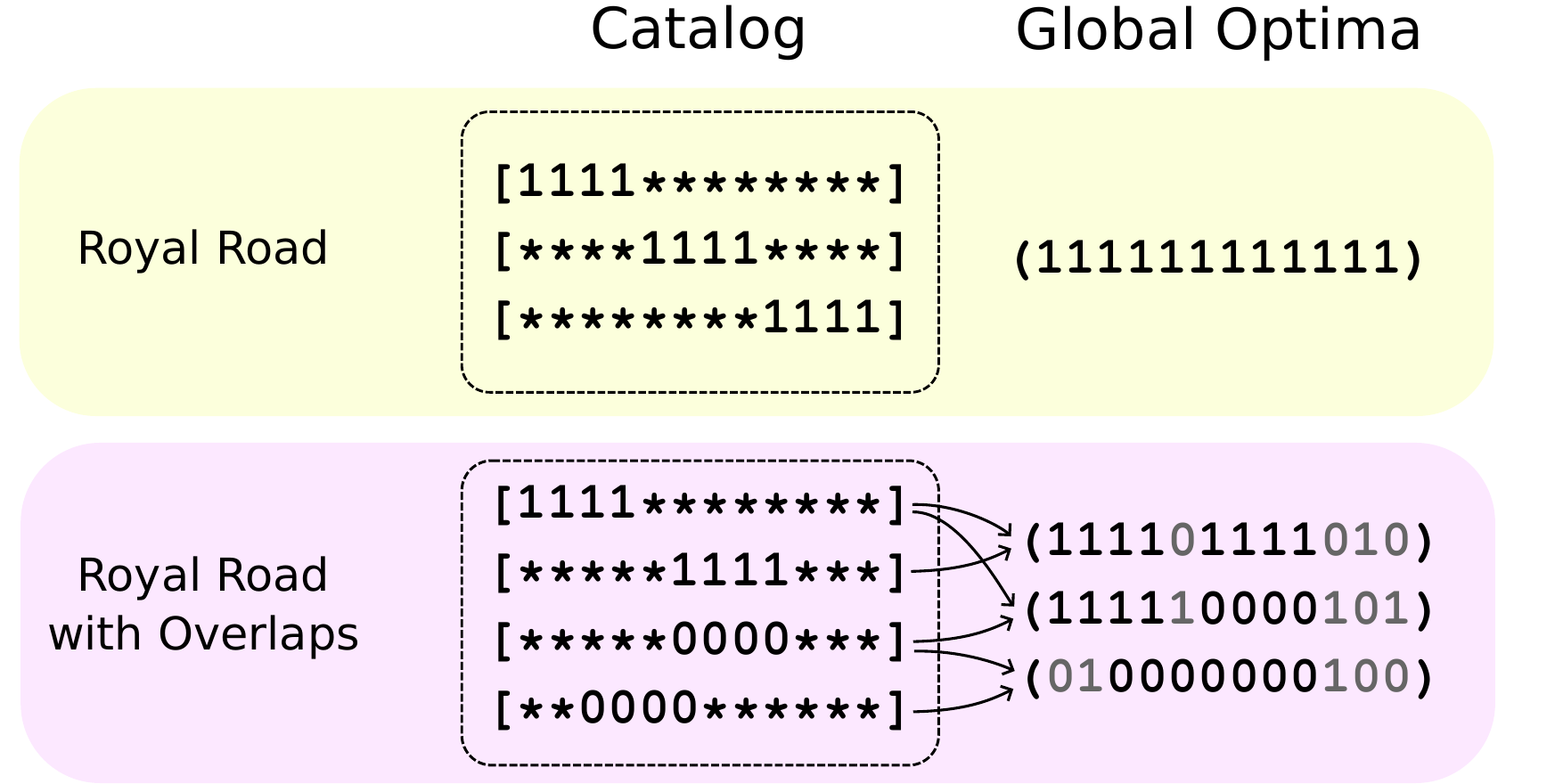}
  \caption{Comparison of the RR and RRO problems, where RRO has more complex global optima}
  \label{fig:RRandRRO}
  \Description{The main difference between the two problems is that the Partial Solutions in RRO can "overlap", meaning that some of them might not be able to coexist in the same full solution. This has the interesting effect of often producing many distinguished global optima, whereas in RR there is only one single global optima, composed of the union of all the items in the PS \PSbasis}
\end{figure}

The \textbf{Royal Road} problem assigns a fitness to the input bit strings by dividing them into disjoint adjacent groups of $k$ variables, and counting how many of them are entirely consisting of 1's \cite{RoyalRoadForGA}.
The target PSs here are the individual groups, with their values all being set to 1. In this study $k = 4$ (as can be seen in \Cref{fig:RRandRRO}), with 20 parameters in total, meaning that there are 5 target PSs.

A property of the Royal Road problem is that the target PSs are disjoint, which might be too trivial for the algorithm to solve. In order to increase the problem complexity, a variant called \textbf{Royal Road with Overlaps} (RRO) will be introduced, in which the groups are allowed to have overlaps and use 0's instead, as seen in \Cref{fig:RRandRRO}.

More formally, RRO is described by $k$ (the size of each group), $q$ (the number of groups) and $l$ (the amount of bits in the total input). 
When $l < q \cdot k$ the groups will be forced to overlap, which increases problem complexity.
Let $G^\psi$ be the targets, in the form of PSs, then the fitness function will be: 
\begin{equation}
    F_{RRO}(x) = |\{g^{\psi} \in G^\psi\ \text{s.t.}\ \text{contains}\ (x, g^{\psi})\}|
\end{equation}

In this paper, $k = 4$, $q = 5$ and $l = 15$ (guaranteeing overlap), with the groups being randomized for every run. A full solution containing all of the groups will often not be possible, and thus the task of finding PSs does not align with finding just the building blocks present in the global optima alone.

Another popular benchmark is \textbf{\TrapK}, in which each group of k bits is assigned a fitness which increases monotonically with the amount of 0's, but achieves its maximum when it contains only 1's, as seen in \Cref{FSubTrap}  \cite{IntroductionToTrapk, OptimalMixingEvolutionaryAlgorithms}.

\begin{align}
    F_{\text{\TrapK}}(x) = \sum_{i=0}^{(l/\kappa)-1} F^{\text{sub}}_{\text{\TrapK}}\left( \sum_{j = \kappa\cdot i}^{(\kappa+1)\cdot i - 1}x_j\right)  \\
    F^{\text{sub}}_{\text{\TrapK}}(u) = 
\begin{cases}
    k & \text{if } u = k, \\
    k - u - 1 & \text{if } u < k.
\end{cases}
\label{FSubTrap}
\end{align}
The problem was devised to be challenging for GAs, due to its deceptive fitness function, and it’s been considered here to check whether deceptive PSs are a problem for our algorithms. Here, $k = 5$ and $l = 25$, so that there will be 5 deceptive groups. The target PSs correspond to these groups, with all their parameters being set to 1.

\subsection{Setup}
\label{subsec:Setup}

\begin{table}[t]
\caption{PS Mining configurations}
\label{table:MinerConfigurations}
\begin{tabular}{ll}
\hline
Algorithm             & Parameters                                                                                                                               \\ \hline
Own method       & $\begin{cases} \text{Population\ Size}\ \in \{50, 100, 150\}\\
                                    \text{Local\ Search}\ \in \begin{Bmatrix}
\begin{matrix}
\text{Simplification-only},\\ 
\text{Specialization-only},\\
\text{Full-local}
\end{matrix}\\

\end{Bmatrix}\\
                                    \text{Uses\ Archive}\ \in \{\text{True}, \text{False}\}
                                \end{cases}$  \\
GA               & $\begin{cases} \text{One point mutation, rate of 0.075} \\ 
                                 \text{Two point crossover, rate of 0.7} \end{cases} $\\
Hill Climber     & $\begin{cases} \text{One point mutation, rate of 0.075}\\ 
                                 \text{Repeated trials, with each trial ending}\\
                                 \text{when a local optima is reached} \end{cases} $\\

\end{tabular}
\end{table}

\begin{table}[t]
\caption{Reference Population GA Parameters}
\label{table:ReferenceGAParameters}
\begin{tabular}{ll}
\hline
Option             &  Value                                                                                                                                 \\ \hline
Selection      & Tournament (size = 2), Elite = 2\\
Mutation & One point mutation, rate of 0.075 \\
Crossover & Two point crossover, rate of 0.7 \\
\end{tabular}
\end{table}

When devising a PS mining algorithm, there are many potential sources of variability, including:
\begin{itemize}
    \item The algorithm itself, due to selection operators
    \item The reference population, being randomly generated
    \item The problem definition, in the case of RRO
\end{itemize}

Due to this, whenever data is gathered about an algorithm configuration being executed on a certain problem, the run is executed 100 times, with a freshly generated reference population (and problem definition for RRO).
Additionally, the algorithms will be restricted to returning at most 50 PSs for every run (\texttt{qty\_ret} in \Cref{algo:PSMiner}).
\subsubsection{T1}
\label{subsubsec:T1}
Many configurations of algorithm parameters will be compared in their ability to \textbf{find PSs} efficiently. This will be done by measuring the number of $\Fpsi$ evaluations needed in order to find all the target PSs for the benchmark problems.
The algorithms and the parameters that will be compared are shown in \Cref{table:MinerConfigurations}.

In each run, $\ReferencePopulation$ will consist of $10^4$ uniformly randomly generated full solutions.
The algorithms will then be executed with the termination criterion being that they have found all of the target PSs, or that they have gone over $10^5$ $\Fpsi$ evaluations. Once the run terminates, it is considered \textbf{successful only if the output \PSbasis contains \textit{all} of the target PSs.}

\subsubsection{T2}
\label{subsubsec:T2}
This  round will contain a short analysis of how the ability to find the PS \PSbasis is affected by the size of $\ReferencePopulation$ and the number of generations it has been evolved for.
Populations of sizes $\in \{100, 200, 500, 1K, 2K, 5K, 10K\}$ are evolved for generations $\in \{0, 10, 20, 50, 100, 200\}$. The evolution process is done using a traditional GA, as described in \Cref{table:ReferenceGAParameters} (derived from \cite{MarkovNetworkSurrogateFitness, MarkovModelCostBenefitAnalysis}), and the \textbf{success criteria is the same as T1.}

\subsubsection{T3}
\label{subsubsec:T3}
The ability for the system to \textbf{produce high-quality full solutions} will be tested by applying \pickAndMerge\ as defined in \Cref{sec:GeneratingFullSolutions}. For evaluation budgets $\in$ $\{$1K, 5K, 10K, 15K, 20K, 25K, 30K$\}$, three methods to generate full solutions will be compared: a traditional \textbf{GA} (as defined for T2), \textbf{UMDA} (both with population size of 150) and the method suggested in \Cref{sec:OverallSystem}.
The latter will use the settings selected from T1 to produce the \PSbasis $\Xpsi$ via the \textbf{PS Miner} algorithm, by consuming a percentage ($\in$ {10\%, 20\%, $\dots$, 90\%}) of the evaluation budget with $\Fpsi$ evaluations. The remaining evaluations are spent on $F$ to evaluate solutions and construct $\ReferencePopulation$. The \PSbasis obtained is passed to \textbf{$\pickAndMerge$} to produce full solutions. 
For each of these three methods, a run is considered \textbf{successful if it produces a global optima for the problem}.

\section{Results}
\label{sec:Results}
\subsection{T1}
\Cref{tab:ResultTableForT1} shows the success rate for each PS Mining algorithm configuration, with the clear winner having the following properties:

\begin{grayBox}
    \begin{itemize}
    \item method: Archive-based
    \item population size: 150
    \item variant: Specialization-only
\end{itemize}
\end{grayBox}

This configuration will be used in the tests that follow.
In general, configurations with the archive enabled were more successful than their counterparts, with the specialization-only variants being the best across all problems. Additionally, the GA and Hill Climber were less successful, particulary with regard to the Trap-k problem. 

% This might indicate that the crossover operator is actually detrimental to the search process in this case, as \PSbasis elements are meant to be atomic (as described in \Cref{sec:FormalDefinition}).

\begin{table}[ht]
\caption{Results for T1: success rates for different algorithm configurations (in bold are figures > 90\%)}
\label{tab:ResultTableForT1}
\centering
\scriptsize
\begin{tabular}{lrrrrrr}\toprule
\multicolumn{3}{c}{\textbf{Search method}} &\multicolumn{3}{c}{\textbf{Problem}} \\\cmidrule{1-6}
\makecell[c]{\textbf{Local search}} &\makecell[c]{\textbf{Pop. size}} &\makecell{\textbf{Uses} \\ \textbf{archive}} &\textbf{RR} &\textbf{RRO} &\textbf{Trap-k} \\
Simplification-only &50 &\multicolumn{1}{l}{\multirow{9}{*}{

$\left\{\begin{matrix}
\\ 
\\ 
\\
\\ 
True\\ 
\\ 
\\ 
\\ 
\\ 

\end{matrix}\right.$}} &0\% &1\% &0\% \\
Simplification-only &100 & &0\% &1\% &0\% \\
Simplification-only &150 & &0\% &5\% &0\% \\
Specialization-only &50 & &64\% &61\% &47\% \\
Specialization-only &100 & &\textbf{96\%} &\textbf{93\%} &86\% \\
Specialization-only &150 & &\textbf{100\%} &\textbf{94\%} &\textbf{93\%} \\
Full-local &50 & &0\% &36\% &0\% \\
Full-local &100 & &0\% &27\% &0\% \\
Full-local &150 & &0\% &31\% &0\% \\
Simplification-only &50 &\multirow{9}{*}{
$\left\{\begin{matrix}
\\ 
\\ 
\\
\\ 
False\\ 
\\ 
\\ 
\\ 
\\ 

\end{matrix}\right.$} &0\% &0\% &0\% \\
Simplification-only &100 & &0\% &0\% &0\% \\
Simplification-only &150 & &0\% &0\% &0\% \\
Specialization-only &50 & &61\% &12\% &1\% \\
Specialization-only &100 & &\textbf{95\%} &85\% &10\% \\
Specialization-only &150 & &\textbf{100\%} &\textbf{93\%} &30\% \\
Full-local &50 & &0\% &0\% &0\% \\
Full-local &100 & &0\% &6\% &0\% \\
Full-local &150 & &0\% &18\% &0\% \\
GA &50 & - &0\% &0\% &0\% \\
GA &100 & -&0\% &0\% &0\% \\
GA &150 & - &0\% &0\% &0\% \\
Hill-climber &- & -&28\% &31\% &0\% \\
\bottomrule
\end{tabular}
\end{table}

\subsection{T2}
The data in \Cref{table:T2Results} suggests that the best $\ReferencePopulation$ to find the \PSbasis is unevolved, with its minimum size = $10^4$ for the tested problems.
A possible explanation for this effect is the phenomenon of ``\textit{hitchhiking}'', where sub-optimal traits spread throughout a population due to their co-occurrence with optimal traits.
This effect was hypothesized by \citet{WhenWillAGAOutperform} upon observing similar results for Royal functions, where methods which rely on schemata (such as \cite{JohnDetrimentalEvolution}) perform worse when applied on pre-evolved populations.

%This effect could also be caused by the definition of the objective functions, especially \texttt{atomicity}, which might be revisited in future research.

\begin{table*}[ht] %table* makes it big and float
\caption{PS \PSbasis success rate for T2, in bold are figures > 90\% (success is finding all of the target Partial Solutions)}
\label{table:T2Results}
%% the magic numbers are -37, .32 and .335
\scriptsize
\begin{tabular}{c}  % puts the tables side by side
\begin{minipage}{.37\linewidth}  % minipage allows individual captioning
\caption*{RR}
\begin{tabular}{lrrrrrrr}\toprule
Gen's &\textbf{0} &\textbf{10} &\textbf{20} &\textbf{50} &\textbf{100} &\textbf{200} \\\midrule
$\ReferencePopulation$ Size & & & & & & & \\\midrule
\textbf{100} &0\% &0\% &0\% &0\% &0\% &0\% \\
\textbf{200} &0\% &0\% &0\% &0\% &0\% &0\% \\
\textbf{500} &52\% &23\% &22\%  &12\% &23\% &23\% \\
\textbf{1000} &\textbf{90\%} &77\% &80\% &75\% &80\% &82\% \\
\textbf{2000} &\textbf{100\%} &\textbf{99\%} &\textbf{98\%}  &\textbf{99\%} &\textbf{98\%} &\textbf{99\%} \\
\textbf{5000} &\textbf{100\%} &\textbf{100\%} &\textbf{100\%} &\textbf{100\%} &\textbf{100\%} &\textbf{100\%} \\
\textbf{10000} &\textbf{100\%} &\textbf{100\%} &\textbf{100\%} &\textbf{100\%} &\textbf{100\%} &\textbf{100\%} \\
\bottomrule
\end{tabular}
\end{minipage}

\begin{minipage}{.32\linewidth} % start of second table
\caption*{RRO}
\begin{tabular}{lrrrrrrr}\toprule
Gen's &\textbf{0} &\textbf{10} &\textbf{20} &\textbf{50} &\textbf{100} &\textbf{200} \\\midrule
$\ReferencePopulation$ Size & & & & & & \\\midrule
\textbf{100} &4\% &0\% &0\% &0\% &0\% &0\% \\
\textbf{200} &22\% &8\% &9\% &4\% &5\% &4\% \\
\textbf{500} &66\% &55\% &47\% &50\% &45\% &46\% \\
\textbf{1000} &83\% &82\% &77\% &75\% &81\% &78\% \\
\textbf{2000} &89\% &88\% &89\% &87\% &87\% &84\% \\
\textbf{5000} &89\% &86\% &88\% &\textbf{91\%} &\textbf{90\%} &89\% \\
\textbf{10000} &\textbf{94\%} &89\% &89\% &89\% &89\% &89\% \\
\bottomrule
\end{tabular}
\end{minipage}

\begin{minipage}{.35\linewidth} % start of third table
\caption*{Trap-k}
\begin{tabular}{lrrrrrrr}\toprule
Gen's &\textbf{0} &\textbf{10} &\textbf{20} &\textbf{50} &\textbf{100} &\textbf{200} \\\midrule
$\ReferencePopulation$ Size & & & & & & \\\midrule
\textbf{100} &0\% &0\% &0\% &0\% &0\% &0\% \\
\textbf{200} &0\% &0\% &0\% &0\% &0\% &0\% \\
\textbf{500} &0\% &0\% &0\% &0\% &0\% &0\% \\
\textbf{1000} &0\% &0\% &0\% &0\% &0\% &0\% \\
\textbf{2000} &2\% &0\% &0\% &0\% &0\% &0\% \\
\textbf{5000} &38\% &11\% &4\% &3\% &3\% &6\% \\
\textbf{10000} &\textbf{93\%} &51\% &60\% &53\% &49\% &60\% \\
\bottomrule
\end{tabular}
\end{minipage}
\end{tabular}

\end{table*}

\begin{figure*}[!ht]
  \centering
  \includegraphics[width=0.9\linewidth]{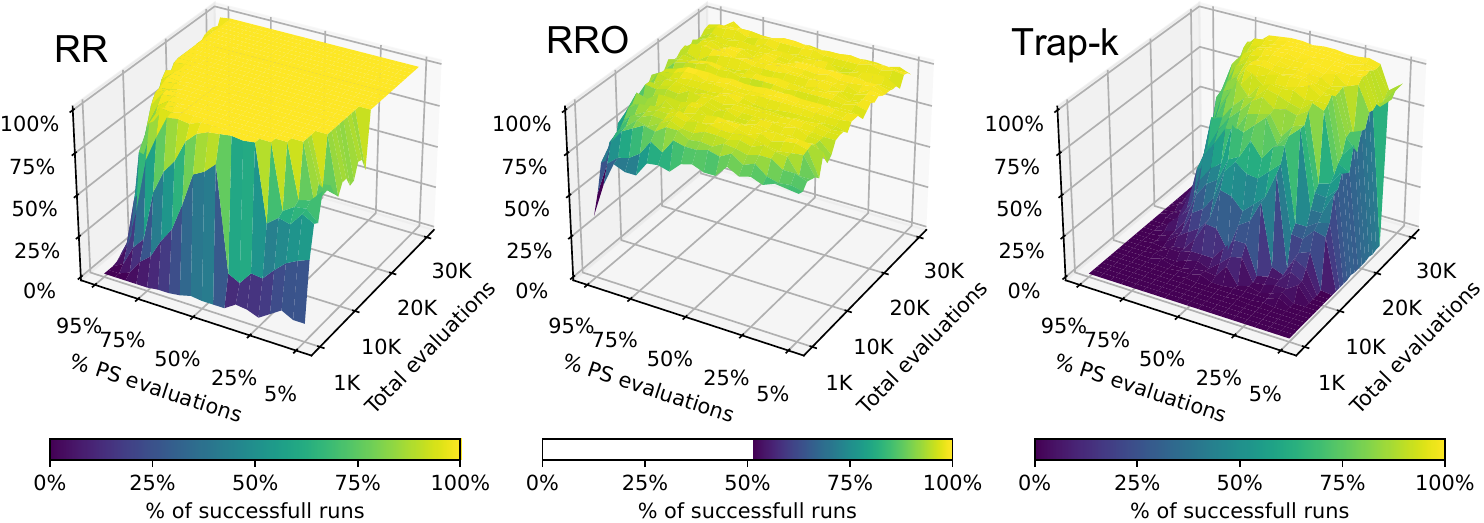}
  \caption{(Relating to T3) Mean fitness obtained by dedicating some of the evaluation budget to PS \PSbasis mining.}
  \label{fig:T3Plot}
  \Description{For all of the problems, the highest mean fitness is obtained by using 20\% to 25\% of the budget to PS evaluations.}
\end{figure*}

\subsection{T3}
\Cref{table:T3Results} and \Cref{fig:T3Plot} show three important facts:
\begin{enumerate}
    \item The PS \PSbasis mediated approach is able to consistently find the global optima for all of the problems, when between 20\% and 60\% being the recommended proportion of the budget to dedicate to $\Fpsi$ evaluations.
    \item for the RR and RRO problems, the GA and the PS \PSbasis mediated approach perform similarly when around 50\% of the evaluations are used for $\Fpsi$.
    \item For Trap-k, only the proposed method was able to find the global optima consistently, whereas the GA and UMDA almost always got stuck in the local optima.
\end{enumerate}

\begin{table*}[ht]
\caption{Global optima success rate for T3, in bold are figures >  90\% (success is finding a global optima)}
\label{table:T3Results}
\scriptsize
\tabcolsep=0.15cm
\begin{tabular}{c}  % puts the tables side by side
\begin{minipage}{.335\linewidth}  % minipage allows individual captioning
\caption*{RR}
\begin{adjustwidth}{-0.3 cm}{-0.3 cm}
\centering
\begin{tabular}{lrrrrrrrrr}\toprule
\multicolumn{2}{c}{Eval's} &1K &5K &10K &15K &20K &25K &30K \\\midrule
\multicolumn{2}{c}{P\&M} & & & & & & & \\
\multirow{9}{*}{\rotatebox[origin=c]{90}{\% of eval's for PS catalog}} &10\% &9\% &71\% &\textbf{100\%} &\textbf{100\%} &\textbf{100\%} &\textbf{100\%} &\textbf{100\%} \\
&20\% &8\% &\textbf{100\%} &\textbf{100\%} &\textbf{100\%} &\textbf{100\%} &\textbf{100\%} &\textbf{100\%} \\
&30\% &6\% &\textbf{100\%} &\textbf{100\%} &\textbf{100\%} &\textbf{100\%} &\textbf{100\%} &\textbf{100\%} \\
&40\% &3\% &\textbf{100\%} &\textbf{100\%} &\textbf{100\%} &\textbf{100\%} &\textbf{100\%} &\textbf{100\%} \\
&50\% &3\% &\textbf{100\%} &\textbf{100\%} &\textbf{100\%} &\textbf{100\%} &\textbf{100\%} &\textbf{100\%} \\
&60\% &2\% &\textbf{100\%} &\textbf{100\%} &\textbf{100\%} &\textbf{100\%} &\textbf{100\%} &\textbf{100\%} \\
&70\% &1\% &\textbf{98\%} &\textbf{100\%} &\textbf{100\%} &\textbf{100\%} &\textbf{100\%} &\textbf{100\%} \\
&80\% &0\% &88\% &\textbf{98\%} &\textbf{100\%} &\textbf{100\%} &\textbf{100\%} &\textbf{100\%} \\
&90\% &0\% &17\% &79\% &\textbf{95\%} &\textbf{100\%} &\textbf{100\%} &\textbf{100\%} \\
\multicolumn{2}{c}{GA} &13\% &\textbf{100\%} &\textbf{100\%} &\textbf{100\%} &\textbf{100\%} &\textbf{100\%} &\textbf{100\%} \\
\multicolumn{2}{c}{UMDA} &0\% &3\% &1\% &0\% &0\% &0\% &0\% \\
\bottomrule
\end{tabular}
\end{adjustwidth}
\end{minipage}

\begin{minipage}{.335\linewidth} % start of second table
\caption*{RRO}
\begin{adjustwidth}{-0.05 cm}{-0.05 cm}
\centering
\begin{tabular}{lrrrrrrrrr}\toprule
\multicolumn{2}{c}{Eval's} &1K &5K &10K &15K &20K &25K &30K \\\midrule
\multicolumn{2}{c}{P\&M} & & & & & & & \\
\multirow{9}{*}{\rotatebox[origin=c]{90}{\% of eval's for PS catalog}} &10\% &86\% &\textbf{96\%} &\textbf{96\%} &\textbf{97\%} &\textbf{98\%} &\textbf{97\%} &\textbf{98\%} \\
&20\% &84\% &\textbf{95\%} &\textbf{97\%} &\textbf{99\%} &\textbf{99\%} &\textbf{98\%} &\textbf{97\%} \\
&30\% &85\% &\textbf{97\%} &\textbf{97\%} &\textbf{98\%} &\textbf{98\%} &\textbf{98\%} &\textbf{96\%} \\
&40\% &81\% &\textbf{97\%} &\textbf{97\%} &\textbf{99\%} &\textbf{96\%} &\textbf{98\%} &\textbf{96\%} \\
&50\% &84\% &\textbf{98\%} &\textbf{98\%} &\textbf{98\%} &\textbf{96\%} &\textbf{97\%} &\textbf{97\%} \\
&60\% &76\% &\textbf{96\%} &\textbf{98\%} &\textbf{99\%} &\textbf{98\%} &\textbf{98\%} &\textbf{97\%} \\
&70\% &77\% &\textbf{94\%} &\textbf{92\%} &\textbf{97\%} &\textbf{96\%} &\textbf{97\%} &\textbf{96\%} \\
&80\% &68\% &\textbf{92\%} &\textbf{93\%} &\textbf{99\%} &\textbf{95\%} &\textbf{96\%} &\textbf{96\%} \\
&90\% &56\% &81\% &84\% &\textbf{96\%} &\textbf{96\%} &\textbf{96\%} &\textbf{96\%} \\
\multicolumn{2}{c}{GA} &\textbf{98\%} &\textbf{100\%} &\textbf{99\%} &\textbf{100\%} &\textbf{99\%} &\textbf{98\%} &\textbf{98\%} \\
\multicolumn{2}{c}{UMDA} &73\% &77\% &69\% &68\% &63\% &73\% &71\% \\
\bottomrule
\end{tabular}
\end{adjustwidth}
\end{minipage}

\begin{minipage}{.335\linewidth} % start of third table
\caption*{Trap-k}
\centering
\begin{tabular}{lrrrrrrrrr}\toprule
\multicolumn{2}{c}{Eval's} &1K &5K &10K &15K &20K &25K &30K \\\midrule
\multicolumn{2}{c}{P\&M} & & & & & & & \\
\multirow{9}{*}{\rotatebox[origin=c]{90}{\% of eval's for PS catalog}} &10\% &0\% &0\% &0\% &53\% &71\% &82\% &81\% \\
&20\% &0\% &0\% &13\% &39\% &\textbf{97\%} &\textbf{99\%} &\textbf{100\%} \\
&30\% &0\% &0\% &5\% &79\% &\textbf{99\%} &\textbf{99\%} &\textbf{99\%} \\
&40\% &0\% &0\% &21\% &87\% &\textbf{95\%} &\textbf{99\%} &\textbf{100\%} \\
&50\% &0\% &0\% &14\% &74\% &\textbf{92\%} &\textbf{99\%} &\textbf{100\%} \\
&60\% &0\% &0\% &6\% &52\% &89\% &\textbf{96\%} &\textbf{99\%} \\
&70\% &0\% &0\% &4\% &38\% &72\% &89\% &\textbf{93\%} \\
&80\% &0\% &0\% &0\% &11\% &24\% &58\% &71\% \\
&90\% &0\% &0\% &0\% &0\% &4\% &3\% &8\% \\
\multicolumn{2}{c}{GA} &0\% &0\% &0\% &0\% &0\% &0\% &0\% \\
\multicolumn{2}{c}{UMDA} &0\% &0\% &0\% &0\% &0\% &0\% &0\% \\
\bottomrule
\end{tabular}
\end{minipage}
\end{tabular}

\end{table*}

These results give evidence that the proposed methodology can solve optimization problems with similar performance to a GA, improving on Trap-k, while adding explainability.

\section{Conclusion}
The algorithms presented in this paper describe a novel technique for explainability, by using Partial Solutions. The PS \PSbasis provides a \textbf{global explanation} of the fitness landscape by presenting beneficial sub-configurations, while \textbf{local explanations} are obtained by checking which PSs are found in a solution.
Moreover, the \PSbasis acts similar to a model in an EDA, from which new solutions can be sampled and thus solving the original optimization problem.
The testing rounds showed the following results:
\begin{itemize}
    \item \textbf{RQ1}: The algorithm which is most apt for finding the PS \PSbasis in the benchmarks was the Specialization-only local search, using the exclusion archive.
    \item \textbf{RQ2}: To obtain the PS \PSbasis, it is better for $\ReferencePopulation$ to be a large unevolved population
    \item \textbf{RQ3}: The system is consistently able to get the global optima for the benchmark problems by first generating a PS \PSbasis and then applying $\pickAndMerge$. This suggests that the PS \PSbasis can capture important information about the fitness landscape. The recommended proportion of 
 $\Fpsi$ evaluations is between 20\% and 60\%.
    
\end{itemize}

It should be noted that the PS Miner algorithms implemented in this work are not as efficient as they could be, since $F^\psi$ involves complex calculations, and experimentation indicates this to be between 50 and 100 times slower than $F$ for $|\ReferencePopulation| = 10^4$.
Therefore the explainability of Partial Solutions still comes at a computational cost, although this is might be partially countered by an improved chance at finding global optima (as seen for Trap-k).
The source code for the system can be viewed at \cite{Catalano_PS_Assisted_Explainability_2024}.

\section{Future work}

The PS \PSbasis represents beneficial traits for the fitness function, and thus they can act as a \textbf{model} that could be used to implement a surrogate fitness function (similar to \cite{SandyThesis}) or improved GA operators (similar to Optimal Mixing in \cite{OptimalMixingEvolutionaryAlgorithms}). 
Overall, the system could be readapted to be a full EDA, for example by feeding the outputs of \pickAndMerge\ into $\ReferencePopulation$.

The problems discussed in this document are admittedly simpler than \textbf{real-world problems}, and therefore it would be useful to see how well the PSs perform with some \textbf{inherently inelegant problems}, such as rostering and scheduling tasks.
In particular, there will be cases where interpretable PSs are not always present or difficult to find. This will be investigated in problems without clean-cut cliques, where the explainability of PSs will be explored more thoroughly.

As mentioned in \Cref{sec:FormalDefinition}, a PS may be \textbf{generalized to higher-order} relationships.
This will both greatly improve the descriptive power of PSs, but also greatly increase the complexity of the search methods.
$\Fpsi$ could also be extended by adding more metrics, such as novelty, robustness (to changes in the parameter values with respect to fitness), or constraint satisfaction.

\begin{acks}
    Work funded under a PhD project supported by The Data Lab and BT Group plc. Many thanks go to the patience of my supervisors.
\end{acks}

%%
%% The next two lines define the bibliography style to be used, and
%% the bibliography file.

%% The following line was removed for submission to ArXiv
\bibliographystyle{ACM-Reference-Format}
\bibliography{bibliography}

%%% -*-BibTeX-*-
%%% Do NOT edit. File created by BibTeX with style
%%% ACM-Reference-Format-Journals [18-Jan-2012].

\begin{thebibliography}{31}

%%% ====================================================================
%%% NOTE TO THE USER: you can override these defaults by providing
%%% customized versions of any of these macros before the \bibliography
%%% command.  Each of them MUST provide its own final punctuation,
%%% except for \shownote{}, \showDOI{}, and \showURL{}.  The latter two
%%% do not use final punctuation, in order to avoid confusing it with
%%% the Web address.
%%%
%%% To suppress output of a particular field, define its macro to expand
%%% to an empty string, or better, \unskip, like this:
%%%
%%% \newcommand{\showDOI}[1]{\unskip}   % LaTeX syntax
%%%
%%% \def \showDOI #1{\unskip}           % plain TeX syntax
%%%
%%% ====================================================================

\ifx \showCODEN    \undefined \def \showCODEN     #1{\unskip}     \fi
\ifx \showDOI      \undefined \def \showDOI       #1{#1}\fi
\ifx \showISBNx    \undefined \def \showISBNx     #1{\unskip}     \fi
\ifx \showISBNxiii \undefined \def \showISBNxiii  #1{\unskip}     \fi
\ifx \showISSN     \undefined \def \showISSN      #1{\unskip}     \fi
\ifx \showLCCN     \undefined \def \showLCCN      #1{\unskip}     \fi
\ifx \shownote     \undefined \def \shownote      #1{#1}          \fi
\ifx \showarticletitle \undefined \def \showarticletitle #1{#1}   \fi
\ifx \showURL      \undefined \def \showURL       {\relax}        \fi
% The following commands are used for tagged output and should be
% invisible to TeX
\providecommand\bibfield[2]{#2}
\providecommand\bibinfo[2]{#2}
\providecommand\natexlab[1]{#1}
\providecommand\showeprint[2][]{arXiv:#2}

\bibitem[\protect\citeauthoryear{Bacardit, Brownlee, Cagnoni, Iacca, McCall, and Walker}{Bacardit et~al\mbox{.}}{2022}]%
        {IntersectionEvoCom}
\bibfield{author}{\bibinfo{person}{Jaume Bacardit}, \bibinfo{person}{Alexander E.~I. Brownlee}, \bibinfo{person}{Stefano Cagnoni}, \bibinfo{person}{Giovanni Iacca}, \bibinfo{person}{John McCall}, {and} \bibinfo{person}{David Walker}.} \bibinfo{year}{2022}\natexlab{}.
\newblock \showarticletitle{The Intersection of Evolutionary Computation and Explainable AI}. In \bibinfo{booktitle}{\emph{Proceedings of the Genetic and Evolutionary Computation Conference Companion}} (Boston, Massachusetts) \emph{(\bibinfo{series}{GECCO '22})}. \bibinfo{publisher}{Association for Computing Machinery}, \bibinfo{address}{New York, NY, USA}, \bibinfo{pages}{1757–1762}.
\newblock
\showISBNx{9781450392686}
\urldef\tempurl%
\url{https://doi.org/10.1145/3520304.3533974}
\showDOI{\tempurl}


\bibitem[\protect\citeauthoryear{Baluja and Davies}{Baluja and Davies}{1997}]%
        {OptimalDependencyTrees}
\bibfield{author}{\bibinfo{person}{Shumeet Baluja} {and} \bibinfo{person}{Scott Davies}.} \bibinfo{year}{1997}\natexlab{}.
\newblock \showarticletitle{Using Optimal Dependency-Trees for Combinational Optimization}. In \bibinfo{booktitle}{\emph{Proceedings of the Fourteenth International Conference on Machine Learning}} \emph{(\bibinfo{series}{ICML '97})}. \bibinfo{publisher}{Morgan Kaufmann Publishers Inc.}, \bibinfo{address}{San Francisco, CA, USA}, \bibinfo{pages}{30–38}.
\newblock
\showISBNx{1558604863}


\bibitem[\protect\citeauthoryear{{Barredo Arrieta}, Díaz-Rodríguez, {Del Ser}, Bennetot, Tabik, Barbado, Garcia, Gil-Lopez, Molina, Benjamins, Chatila, and Herrera}{{Barredo Arrieta} et~al\mbox{.}}{2020}]%
        {XAIConcepts}
\bibfield{author}{\bibinfo{person}{Alejandro {Barredo Arrieta}}, \bibinfo{person}{Natalia Díaz-Rodríguez}, \bibinfo{person}{Javier {Del Ser}}, \bibinfo{person}{Adrien Bennetot}, \bibinfo{person}{Siham Tabik}, \bibinfo{person}{Alberto Barbado}, \bibinfo{person}{Salvador Garcia}, \bibinfo{person}{Sergio Gil-Lopez}, \bibinfo{person}{Daniel Molina}, \bibinfo{person}{Richard Benjamins}, \bibinfo{person}{Raja Chatila}, {and} \bibinfo{person}{Francisco Herrera}.} \bibinfo{year}{2020}\natexlab{}.
\newblock \showarticletitle{Explainable Artificial Intelligence (XAI): Concepts, taxonomies, opportunities and challenges toward responsible AI}.
\newblock \bibinfo{journal}{\emph{Information Fusion}}  \bibinfo{volume}{58} (\bibinfo{year}{2020}), \bibinfo{pages}{82--115}.
\newblock
\showISSN{1566-2535}
\urldef\tempurl%
\url{https://doi.org/10.1016/j.inffus.2019.12.012}
\showDOI{\tempurl}


\bibitem[\protect\citeauthoryear{Brown, Garmendia-Doval, and McCall}{Brown et~al\mbox{.}}{2002}]%
        {JohnDetrimentalEvolution}
\bibfield{author}{\bibinfo{person}{D.F. Brown}, \bibinfo{person}{A.B. Garmendia-Doval}, {and} \bibinfo{person}{J.A.W. McCall}.} \bibinfo{year}{2002}\natexlab{}.
\newblock \showarticletitle{Markov Random Field Modelling of Royal Road Genetic Algorithms}. In \bibinfo{booktitle}{\emph{Artificial Evolution}}, \bibfield{editor}{\bibinfo{person}{Pierre Collet}, \bibinfo{person}{Cyril Fonlupt}, \bibinfo{person}{Jin-Kao Hao}, \bibinfo{person}{Evelyne Lutton}, {and} \bibinfo{person}{Marc Schoenauer}} (Eds.). \bibinfo{publisher}{Springer Berlin Heidelberg}, \bibinfo{address}{Berlin, Heidelberg}, \bibinfo{pages}{65--76}.
\newblock
\showISBNx{978-3-540-46033-6}


\bibitem[\protect\citeauthoryear{Brownlee}{Brownlee}{2016}]%
        {MarkovNetoworkValueAdded}
\bibfield{author}{\bibinfo{person}{Alexander~E.I. Brownlee}.} \bibinfo{year}{2016}\natexlab{}.
\newblock \showarticletitle{Mining Markov Network Surrogates for Value-Added Optimisation}. In \bibinfo{booktitle}{\emph{Proceedings of the 2016 on Genetic and Evolutionary Computation Conference Companion}} (Denver, Colorado, USA) \emph{(\bibinfo{series}{GECCO '16 Companion})}. \bibinfo{publisher}{Association for Computing Machinery}, \bibinfo{address}{New York, NY, USA}, \bibinfo{pages}{1267–1274}.
\newblock
\showISBNx{9781450343237}
\urldef\tempurl%
\url{https://doi.org/10.1145/2908961.2931711}
\showDOI{\tempurl}


\bibitem[\protect\citeauthoryear{Brownlee}{Brownlee}{2009}]%
        {SandyThesis}
\bibfield{author}{\bibinfo{person}{Alexander Edward~Ian Brownlee}.} \bibinfo{year}{2009}\natexlab{}.
\newblock \emph{\bibinfo{title}{Multivariate Markov networks for fitness modelling in an estimation of distribution algorithm.}}
\newblock \bibinfo{thesistype}{Ph.\,D. Dissertation}. \bibinfo{school}{"Robert Gordon University"}.
\newblock


\bibitem[\protect\citeauthoryear{Brownlee, Regnier-Coudert, McCall, and Massie}{Brownlee et~al\mbox{.}}{2010}]%
        {MarkovNetworkSurrogateFitness}
\bibfield{author}{\bibinfo{person}{Alexander E.~I. Brownlee}, \bibinfo{person}{Olivier Regnier-Coudert}, \bibinfo{person}{John A.~W. McCall}, {and} \bibinfo{person}{Stewart Massie}.} \bibinfo{year}{2010}\natexlab{}.
\newblock \showarticletitle{Using a Markov network as a surrogate fitness function in a genetic algorithm}. In \bibinfo{booktitle}{\emph{IEEE Congress on Evolutionary Computation}}. \bibinfo{publisher}{IEEE}, \bibinfo{address}{New York City, USA}, \bibinfo{pages}{1--8}.
\newblock
\urldef\tempurl%
\url{https://doi.org/10.1109/CEC.2010.5586548}
\showDOI{\tempurl}


\bibitem[\protect\citeauthoryear{Catalano}{Catalano}{2024}]%
        {Catalano_PS_Assisted_Explainability_2024}
\bibfield{author}{\bibinfo{person}{Giancarlo Catalano}.} \bibinfo{year}{2024}\natexlab{}.
\newblock \bibinfo{booktitle}{\emph{{PS Assisted Explainability}}}.
\newblock
\urldef\tempurl%
\url{https://github.com/Giancarlo-Catalano/PS_Minimal_Showcase}
\showURL{%
\tempurl}


\bibitem[\protect\citeauthoryear{Fletcher and Wennekers}{Fletcher and Wennekers}{2017}]%
        {NaturalApproachSchemaProcessing}
\bibfield{author}{\bibinfo{person}{Jack~McKay Fletcher} {and} \bibinfo{person}{Thomas Wennekers}.} \bibinfo{year}{2017}\natexlab{}.
\newblock \bibinfo{title}{A natural approach to studying schema processing}.
\newblock
\newblock
\showeprint[arxiv]{1705.04536}~[cs.NE]


\bibitem[\protect\citeauthoryear{Fyvie, McCall, and Christie}{Fyvie et~al\mbox{.}}{2021a}]%
        {NDSXAITrajectory}
\bibfield{author}{\bibinfo{person}{Martin Fyvie}, \bibinfo{person}{John~A.W. McCall}, {and} \bibinfo{person}{Lee~A. Christie}.} \bibinfo{year}{2021}\natexlab{a}.
\newblock \showarticletitle{Non-deterministic solvers and explainable AI through trajectory mining.}
\newblock \bibinfo{journal}{\emph{SICSA XAI Workshop 2021}}  \bibinfo{volume}{2894} (\bibinfo{date}{6} \bibinfo{year}{2021}), \bibinfo{pages}{75--78}.
\newblock
\showISSN{1613-0073}
\urldef\tempurl%
\url{https://rgu-repository.worktribe.com/output/1395881 https://rgu-repository.worktribe.com/output/1395881.abstract}
\showURL{%
\tempurl}


\bibitem[\protect\citeauthoryear{Fyvie, McCall, Christie, Brownlee, and Singh}{Fyvie et~al\mbox{.}}{2021b}]%
        {PCATrajectoryMining}
\bibfield{author}{\bibinfo{person}{Martin Fyvie}, \bibinfo{person}{John A.~W. McCall}, \bibinfo{person}{Lee~A. Christie}, \bibinfo{person}{Alexander E.~I. Brownlee}, {and} \bibinfo{person}{Manjinder Singh}.} \bibinfo{year}{2021}\natexlab{b}.
\newblock \showarticletitle{Towards explainable metaheuristics: Feature extraction from trajectory mining}.
\newblock \bibinfo{journal}{\emph{Expert Systems}} \bibinfo{volume}{n/a}, \bibinfo{number}{n/a} (\bibinfo{year}{2021}), \bibinfo{pages}{e13494}.
\newblock
\urldef\tempurl%
\url{https://doi.org/10.1111/exsy.13494}
\showDOI{\tempurl}
\showeprint{https://onlinelibrary.wiley.com/doi/pdf/10.1111/exsy.13494}


\bibitem[\protect\citeauthoryear{Goldberg}{Goldberg}{1989}]%
        {IntroductionToTrapk}
\bibfield{author}{\bibinfo{person}{David~E Goldberg}.} \bibinfo{year}{1989}\natexlab{}.
\newblock \showarticletitle{Genetic algorithms and Walsh functions: Part 2, Deception and its analysis}.
\newblock \bibinfo{journal}{\emph{Complex systems}}  \bibinfo{volume}{3} (\bibinfo{year}{1989}), \bibinfo{pages}{153--171}.
\newblock


\bibitem[\protect\citeauthoryear{Goldberg, Sastry, and Ohsawa}{Goldberg et~al\mbox{.}}{2003}]%
        {DiscoveringDeepBuildingBlocks}
\bibfield{author}{\bibinfo{person}{David~E. Goldberg}, \bibinfo{person}{Kumara Sastry}, {and} \bibinfo{person}{Yukio Ohsawa}.} \bibinfo{year}{2003}\natexlab{}.
\newblock \bibinfo{booktitle}{\emph{Discovering Deep Building Blocks for Competent Genetic Algorithms Using Chance Discovery via KeyGraphs}}.
\newblock \bibinfo{publisher}{Springer, Berlin, Heidelberg}, \bibinfo{address}{Tokyo, Japan}, \bibinfo{pages}{276--301}.
\newblock
\showISBNx{978-3-662-06230-2}
\urldef\tempurl%
\url{https://doi.org/10.1007/978-3-662-06230-2_19}
\showDOI{\tempurl}


\bibitem[\protect\citeauthoryear{Grace, Salvatier, Dafoe, Zhang, and Evans}{Grace et~al\mbox{.}}{2018}]%
        {AIExceedHuman}
\bibfield{author}{\bibinfo{person}{Katja Grace}, \bibinfo{person}{John Salvatier}, \bibinfo{person}{Allan Dafoe}, \bibinfo{person}{Baobao Zhang}, {and} \bibinfo{person}{Owain Evans}.} \bibinfo{year}{2018}\natexlab{}.
\newblock \showarticletitle{Viewpoint: When Will AI Exceed Human Performance? Evidence from AI Experts}.
\newblock \bibinfo{journal}{\emph{Journal of Artificial Intelligence Research}}  \bibinfo{volume}{62} (\bibinfo{date}{7} \bibinfo{year}{2018}), \bibinfo{pages}{729--754}.
\newblock
\showISSN{1076-9757}
\urldef\tempurl%
\url{https://doi.org/10.1613/JAIR.1.11222}
\showDOI{\tempurl}


\bibitem[\protect\citeauthoryear{Hartmanis}{Hartmanis}{2006}]%
        {GuideToNP}
\bibfield{author}{\bibinfo{person}{Juris Hartmanis}.} \bibinfo{year}{2006}\natexlab{}.
\newblock \showarticletitle{Computers and Intractability: A Guide to the Theory of NP-Completeness (Michael R. Garey and David S. Johnson)}.
\newblock \bibinfo{journal}{\emph{https://doi.org/10.1137/1024022}}  \bibinfo{volume}{24} (\bibinfo{date}{7} \bibinfo{year}{2006}), \bibinfo{pages}{90--91}.
\newblock
Issue 1.
\showISSN{0036-1445}
\urldef\tempurl%
\url{https://doi.org/10.1137/1024022}
\showDOI{\tempurl}


\bibitem[\protect\citeauthoryear{Holland}{Holland}{1992}]%
        {HollandAdaptation}
\bibfield{author}{\bibinfo{person}{John~H. Holland}.} \bibinfo{year}{1992}\natexlab{}.
\newblock \bibinfo{booktitle}{\emph{{Adaptation in Natural and Artificial Systems: An Introductory Analysis with Applications to Biology, Control, and Artificial Intelligence}}}.
\newblock \bibinfo{publisher}{The MIT Press}, \bibinfo{address}{Massachussets, USA}.
\newblock
\showISBNx{9780262275552}
\urldef\tempurl%
\url{https://doi.org/10.7551/mitpress/1090.001.0001}
\showDOI{\tempurl}


\bibitem[\protect\citeauthoryear{Hsu and Yu}{Hsu and Yu}{2015}]%
        {DSMGAII}
\bibfield{author}{\bibinfo{person}{Shih-Huan Hsu} {and} \bibinfo{person}{Tian-Li Yu}.} \bibinfo{year}{2015}\natexlab{}.
\newblock \showarticletitle{Optimization by pairwise linkage detection, incremental linkage set, and restricted/back mixing: DSMGA-II}. In \bibinfo{booktitle}{\emph{Proceedings of the 2015 Annual Conference on Genetic and Evolutionary Computation}}. \bibinfo{publisher}{Association for Computing Machinery}, \bibinfo{address}{New York, NY, USA}, \bibinfo{pages}{519--526}.
\newblock


\bibitem[\protect\citeauthoryear{Mitchell, Forrest, and Holland}{Mitchell et~al\mbox{.}}{1992}]%
        {RoyalRoadForGA}
\bibfield{author}{\bibinfo{person}{Melanie Mitchell}, \bibinfo{person}{Stephanie Forrest}, {and} \bibinfo{person}{John Holland}.} \bibinfo{year}{1992}\natexlab{}.
\newblock \showarticletitle{The Royal Road for Genetic Algorithms: Fitness Landscapes and GA Performance}.
\newblock \bibinfo{journal}{\emph{European Conference on Artificial Life}}  \bibinfo{volume}{1} (\bibinfo{date}{11} \bibinfo{year}{1992}), \bibinfo{pages}{23--33}.
\newblock


\bibitem[\protect\citeauthoryear{Mitchell, Holland, and Forrest}{Mitchell et~al\mbox{.}}{1993}]%
        {WhenWillAGAOutperform}
\bibfield{author}{\bibinfo{person}{Melanie Mitchell}, \bibinfo{person}{John~H. Holland}, {and} \bibinfo{person}{Stephanie Forrest}.} \bibinfo{year}{1993}\natexlab{}.
\newblock \showarticletitle{When will a genetic algorithm outperform hill climbing?}. In \bibinfo{booktitle}{\emph{Proceedings of the 6th International Conference on Neural Information Processing Systems}} (Denver, Colorado) \emph{(\bibinfo{series}{NIPS'93})}. \bibinfo{publisher}{Morgan Kaufmann Publishers Inc.}, \bibinfo{address}{San Francisco, CA, USA}, \bibinfo{pages}{51–58}.
\newblock


\bibitem[\protect\citeauthoryear{Oh and Ahn}{Oh and Ahn}{2023}]%
        {InterpretableFeatureSelectionManifacturing}
\bibfield{author}{\bibinfo{person}{Sanghoun Oh} {and} \bibinfo{person}{Chang~Wook Ahn}.} \bibinfo{year}{2023}\natexlab{}.
\newblock \showarticletitle{Evolutionary Approach for Interpretable Feature Selection Algorithm in Manufacturing Industry}.
\newblock \bibinfo{journal}{\emph{IEEE Access}}  \bibinfo{volume}{11} (\bibinfo{year}{2023}), \bibinfo{pages}{46604--46614}.
\newblock
\showISSN{21693536}
\urldef\tempurl%
\url{https://doi.org/10.1109/ACCESS.2023.3274490}
\showDOI{\tempurl}


\bibitem[\protect\citeauthoryear{Pelikan, Goldberg, and Lobo}{Pelikan et~al\mbox{.}}{2002}]%
        {SurveyOptimizationBuilding}
\bibfield{author}{\bibinfo{person}{Martin Pelikan}, \bibinfo{person}{David~E Goldberg}, {and} \bibinfo{person}{Fernando~G Lobo}.} \bibinfo{year}{2002}\natexlab{}.
\newblock \showarticletitle{A survey of optimization by building and using probabilistic models}.
\newblock \bibinfo{journal}{\emph{Computational optimization and applications}}  \bibinfo{volume}{21} (\bibinfo{year}{2002}), \bibinfo{pages}{5--20}.
\newblock


\bibitem[\protect\citeauthoryear{Rabiner}{Rabiner}{1984}]%
        {CombinatorialOptimization}
\bibfield{author}{\bibinfo{person}{L. Rabiner}.} \bibinfo{year}{1984}\natexlab{}.
\newblock \showarticletitle{Combinatorial optimization:Algorithms and complexity}.
\newblock \bibinfo{journal}{\emph{IEEE Transactions on Acoustics, Speech, and Signal Processing}}  \bibinfo{volume}{32} (\bibinfo{date}{12} \bibinfo{year}{1984}), \bibinfo{pages}{1258--1259}.
\newblock
Issue 6.
\showISSN{0096-3518}
\urldef\tempurl%
\url{https://doi.org/10.1109/TASSP.1984.1164450}
\showDOI{\tempurl}


\bibitem[\protect\citeauthoryear{Shakya, McCall, and Brown}{Shakya et~al\mbox{.}}{2005}]%
        {MarkovModelCostBenefitAnalysis}
\bibfield{author}{\bibinfo{person}{Siddhartha Shakya}, \bibinfo{person}{John McCall}, {and} \bibinfo{person}{Deryck Brown}.} \bibinfo{year}{2005}\natexlab{}.
\newblock \showarticletitle{Using a Markov Network Model in a Univariate EDA: An Empirical Cost-Benefit Analysis}. In \bibinfo{booktitle}{\emph{Proceedings of the 7th Annual Conference on Genetic and Evolutionary Computation}} (Washington DC, USA) \emph{(\bibinfo{series}{GECCO '05})}. \bibinfo{publisher}{Association for Computing Machinery}, \bibinfo{address}{New York, NY, USA}, \bibinfo{pages}{727–734}.
\newblock
\showISBNx{1595930108}
\urldef\tempurl%
\url{https://doi.org/10.1145/1068009.1068130}
\showDOI{\tempurl}


\bibitem[\protect\citeauthoryear{Slaney and Walsh}{Slaney and Walsh}{2001}]%
        {BackbonesInOptimization}
\bibfield{author}{\bibinfo{person}{John Slaney} {and} \bibinfo{person}{Toby Walsh}.} \bibinfo{year}{2001}\natexlab{}.
\newblock \showarticletitle{Backbones in Optimization and Approximation}. In \bibinfo{booktitle}{\emph{Proceedings of the 17th International Joint Conference on Artificial Intelligence - Volume 1}} (Seattle, WA, USA) \emph{(\bibinfo{series}{IJCAI'01})}. \bibinfo{publisher}{Morgan Kaufmann Publishers Inc.}, \bibinfo{address}{San Francisco, CA, USA}, \bibinfo{pages}{254–259}.
\newblock
\showISBNx{1558608125}
\urldef\tempurl%
\url{https://doi.org/10.5555/1642090.1642125}
\showDOI{\tempurl}


\bibitem[\protect\citeauthoryear{Thierens and Bosman}{Thierens and Bosman}{2011a}]%
        {GOMEA}
\bibfield{author}{\bibinfo{person}{Dirk Thierens} {and} \bibinfo{person}{Peter~AN Bosman}.} \bibinfo{year}{2011}\natexlab{a}.
\newblock \showarticletitle{Optimal mixing evolutionary algorithms}. In \bibinfo{booktitle}{\emph{Proceedings of the 13th annual conference on Genetic and evolutionary computation}}. \bibinfo{publisher}{Association for Computing Machinery}, \bibinfo{address}{New York, NY, USA}, \bibinfo{pages}{617--624}.
\newblock


\bibitem[\protect\citeauthoryear{Thierens and Bosman}{Thierens and Bosman}{2011b}]%
        {OptimalMixingEvolutionaryAlgorithms}
\bibfield{author}{\bibinfo{person}{Dirk Thierens} {and} \bibinfo{person}{Peter~A.N. Bosman}.} \bibinfo{year}{2011}\natexlab{b}.
\newblock \showarticletitle{Optimal Mixing Evolutionary Algorithms}. In \bibinfo{booktitle}{\emph{Proceedings of the 13th Annual Conference on Genetic and Evolutionary Computation}} (Dublin, Ireland) \emph{(\bibinfo{series}{GECCO '11})}. \bibinfo{publisher}{Association for Computing Machinery}, \bibinfo{address}{New York, NY, USA}, \bibinfo{pages}{617–624}.
\newblock
\showISBNx{9781450305570}
\urldef\tempurl%
\url{https://doi.org/10.1145/2001576.2001661}
\showDOI{\tempurl}


\bibitem[\protect\citeauthoryear{Thornton}{Thornton}{1997}]%
        {BuildingBlockFallacy}
\bibfield{author}{\bibinfo{person}{Chris Thornton}.} \bibinfo{year}{1997}\natexlab{}.
\newblock \showarticletitle{The building block fallacy}.
\newblock \bibinfo{journal}{\emph{Complexity International}}  \bibinfo{volume}{4} (\bibinfo{year}{1997}), \bibinfo{pages}{1--1}.
\newblock
\showISSN{13200682}


\bibitem[\protect\citeauthoryear{Vance}{Vance}{2006}]%
        {KnapsackProblems}
\bibfield{author}{\bibinfo{person}{Pamela~H. Vance}.} \bibinfo{year}{2006}\natexlab{}.
\newblock \showarticletitle{Knapsack Problems: Algorithms and Computer Implementations (S. Martello and P. Toth)}.
\newblock \bibinfo{journal}{\emph{https://doi.org/10.1137/1035174}}  \bibinfo{volume}{35} (\bibinfo{date}{7} \bibinfo{year}{2006}), \bibinfo{pages}{684--685}.
\newblock
Issue 4.
\showISSN{0036-1445}
\urldef\tempurl%
\url{https://doi.org/10.1137/1035174}
\showDOI{\tempurl}


\bibitem[\protect\citeauthoryear{White}{White}{2014}]%
        {OverviewSchemaTheory}
\bibfield{author}{\bibinfo{person}{David White}.} \bibinfo{year}{2014}\natexlab{}.
\newblock \bibinfo{title}{An Overview of Schema Theory}.
\newblock
\newblock
\showeprint[arxiv]{1401.2651}~[cs.NE]


\bibitem[\protect\citeauthoryear{Xu, Uszkoreit, Du, Fan, Zhao, and Zhu}{Xu et~al\mbox{.}}{2019}]%
        {XAISurvey}
\bibfield{author}{\bibinfo{person}{Feiyu Xu}, \bibinfo{person}{Hans Uszkoreit}, \bibinfo{person}{Yangzhou Du}, \bibinfo{person}{Wei Fan}, \bibinfo{person}{Dongyan Zhao}, {and} \bibinfo{person}{Jun Zhu}.} \bibinfo{year}{2019}\natexlab{}.
\newblock \showarticletitle{Explainable AI: A Brief Survey on History, Research Areas, Approaches and Challenges}.
\newblock \bibinfo{journal}{\emph{Lecture Notes in Computer Science (including subseries Lecture Notes in Artificial Intelligence and Lecture Notes in Bioinformatics)}}  \bibinfo{volume}{11839 LNAI} (\bibinfo{year}{2019}), \bibinfo{pages}{563--574}.
\newblock
\showISBNx{9783030322359}
\showISSN{16113349}
\urldef\tempurl%
\url{https://doi.org/10.1007/978-3-030-32236-6_51/FIGURES/12}
\showDOI{\tempurl}


\bibitem[\protect\citeauthoryear{Yu and Goldberg}{Yu and Goldberg}{2006}]%
        {ConqueringHierarchicalDifficulty}
\bibfield{author}{\bibinfo{person}{Tian-Li Yu} {and} \bibinfo{person}{David~E. Goldberg}.} \bibinfo{year}{2006}\natexlab{}.
\newblock \showarticletitle{Conquering Hierarchical Difficulty by Explicit Chunking: Substructural Chromosome Compression}. In \bibinfo{booktitle}{\emph{Proceedings of the 8th Annual Conference on Genetic and Evolutionary Computation}} (Seattle, Washington, USA) \emph{(\bibinfo{series}{GECCO '06})}. \bibinfo{publisher}{Association for Computing Machinery}, \bibinfo{address}{New York, NY, USA}, \bibinfo{pages}{1385–1392}.
\newblock
\showISBNx{1595931864}
\urldef\tempurl%
\url{https://doi.org/10.1145/1143997.1144210}
\showDOI{\tempurl}


\end{thebibliography}

\end{document}